\documentclass[conference]{IEEEtran}
\IEEEoverridecommandlockouts
\usepackage{cite}
\usepackage{amsmath,amssymb,amsfonts}
\usepackage{algorithmic}
\usepackage{graphicx}
\usepackage{textcomp}
\usepackage{xcolor}
\usepackage{booktabs}
\usepackage{colortbl}
\usepackage{CJKutf8}
\usepackage{tcolorbox}
\usepackage{tabularx}
\usepackage{adjustbox}

\usepackage[colorlinks,linkcolor=blue,anchorcolor=blue,citecolor=blue]{hyperref}
\def\BibTeX{{\rm B\kern-.05em{\sc i\kern-.025em b}\kern-.08em
    T\kern-.1667em\lower.7ex\hbox{E}\kern-.125emX}}
\begin{document}

\title{PsyChat: A Client-Centric Dialogue System for Mental Health Support}

\author{\IEEEauthorblockN{Huachuan Qiu$^{1, 2}$, Anqi Li$^{1, 2}$, Lizhi Ma$^{2}$, Zhenzhong Lan$^{2, \dagger}$\thanks{$^{\dagger}$ Corresponding Author.}}
\IEEEauthorblockA{
$^1$\textit{Zhejiang University}, Hangzhou, China \\
$^2$\textit{School of Engineering, Westlake University}, Hangzhou, China \\
\{qiuhuachuan, lanzhenzhong\}@westlake.edu.cn}
}



\maketitle

\begin{abstract}
Dialogue systems are increasingly integrated into mental health support to help clients facilitate exploration, gain insight, take action, and ultimately heal themselves. A practical and user-friendly dialogue system should be client-centric, focusing on the client's behaviors. However, existing dialogue systems publicly available for mental health support often concentrate solely on the counselor's strategies rather than the behaviors expressed by clients. This can lead to unreasonable or inappropriate counseling strategies and corresponding responses generated by the dialogue system. To address this issue, we propose PsyChat, a client-centric dialogue system that provides psychological support through online chat. The client-centric dialogue system comprises five modules: \textit{client behavior recognition}, \textit{counselor strategy selection}, \textit{input packer}, \textit{response generator}, and \textit{response selection}. Both automatic and human evaluations demonstrate the effectiveness and practicality of our proposed dialogue system for real-life mental health support. Furthermore, the case study demonstrates that the dialogue system can predict the client's behaviors, select appropriate counselor strategies, and generate accurate and suitable responses.

\end{abstract}


\section{Introduction}
Mental health \cite{prince2007no} is a growing concern in our fast-paced and digitally connected world. However, traditional mental health support services often face challenges related to accessibility, affordability, and stigma. Additionally, face-to-face interviews with counselors are constrained by time and space. Therefore, many individuals are hesitant to seek help due to these barriers, leaving their mental well-being at risk. With the increasing demand for mental health support, there is a pressing need for innovative approaches to effectively meet this demand.

\begin{figure}[t!]
    \centering
    \includegraphics[width=8.6cm]{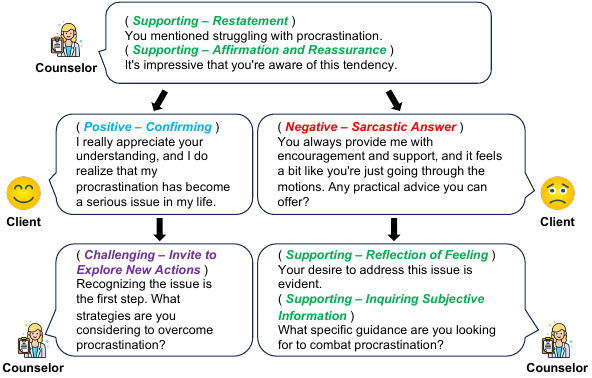}
    \caption{An illustration depicting how a counselor tailors strategies in response to the behaviors exhibited by the client.}
    \label{fig:demo}
\end{figure}

Dialogue systems are increasingly integrated into mental health support to assist clients in exploring, gaining insight, taking action, and ultimately facilitating self-healing \cite{hill2009helping}. A practical and user-friendly dialogue system should be client-centric, focusing on clients' behaviors. However, existing dialogue systems \cite{sun2021psyqa,liu2021towards,zheng2022augesc} for mental health support often concentrate solely on counselors' strategies, frequently overlooking the behaviors expressed by clients. This tendency leads to unreasonable or inappropriate counseling strategies and corresponding responses produced by the dialogue system. More specifically, a practical and user-friendly dialogue system should prioritize considering clients' states. Therefore, it should adjust its strategies based on clients' current behaviors, mimicking human counselors, as illustrated in Figure \ref{fig:demo}.

In light of this, we introduce \textbf{PsyChat}, a client-centric dialogue system designed to provide \underline{psy}chological support through online \underline{chat}. The client-centric dialogue system consists of five modules: client behavior recognition, counselor strategy selection, input packer, response generator, and response selection. The response generation module is intentionally fine-tuned using synthetic and real-life dialogue datasets. The primary contributions of this paper are as follows:

\begin{itemize}
    \item To the best of our knowledge, we are the first to propose a client-centric dialogue system for mental health support, with a priority on considering the client's behaviors.
    \item We optimize collaboration among modules by conducting extensive experiments to identify the optimal model for each. These selected models are then integrated to form a cohesive dialogue system dedicated to mental health.
    \item Automatic and human evaluations demonstrate the effectiveness and practicality of our developed dialogue system. Finally, we release our code and model\footnote{\url{https://github.com/qiuhuachuan/PsyChat}} to facilitate research in mental health support.
\end{itemize}

\begin{figure*}[t!]
    \centering
    \includegraphics[width=0.9\textwidth]{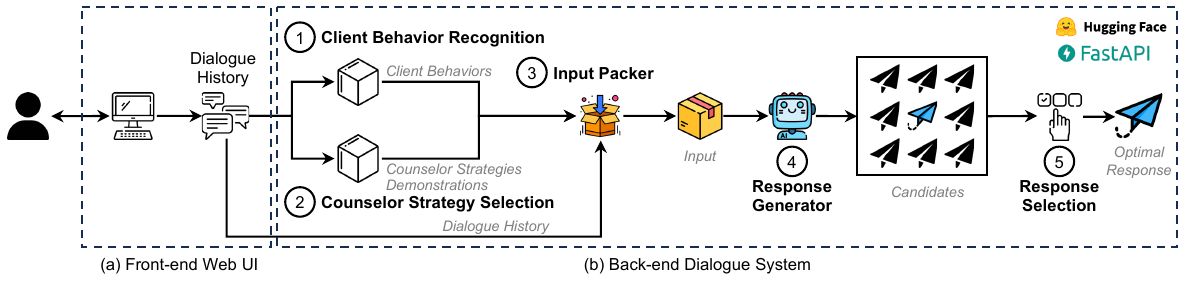}
    \caption{Architecture of the client-centric dialogue system. It contains two main parts: (a) Front-end web UI and (b) Back-end dialogue system, which is implemented with Huggingface and FastAPI packages.}
    \label{fig:architecture}
\end{figure*}

\section{Related Work}\label{sec:related-work}
While the development of integrated dialogue systems for mental health support remains unexplored, we offer a summary of related efforts in constructing a dialogue system for mental health support, examining it from the perspectives of dataset and dialogue model. Importantly, the dataset serves as the cornerstone for building a dialogue model.

\paragraph{Dataset}
Because of privacy concerns in mental health, most dialogue datasets for mental health support are sourced from public social platforms, crowdsourcing, and data synthesis. Dialogue datasets collected from public social platforms include the PsyQA dataset \cite{sun2021psyqa}. Crowdsourcing involves high costs and time, with the ESConv dataset \cite{liu2021towards} serving as a typical example. Data synthesis is an effective approach in the era of large language models, often yielding a large-scale corpus. Some typical datasets include AugESC \cite{zheng2022augesc}, SoulChat \cite{chen2023soulchat}, and SmileChat \cite{qiu2023smile}. Fortunately, a real-world counseling dataset named Xinling \cite{li2023understanding} is conditionally open-sourced, requiring interested users to sign a data usage agreement before utilizing it.

\paragraph{Dialogue Model}
Open-source dialogue models for mental health support contribute novel additions to the research community, including MeChat \cite{qiu2023smile}, SoulChat \cite{chen2023soulchat}, and ChatCounselor \cite{liu2023chatcounselor}.

\section{A Client-Centric Dialogue System for Mental Health Support}
The detailed architecture of the client-centric dialogue system for mental health support is illustrated in Figure \ref{fig:architecture}, which comprises five modules: \textbf{client behavior recognition} ($\S$\ref{cbr}), \textbf{counselor strategy selection} ($\S$\ref{css}), \textbf{input packer} ($\S$\ref{ip}), \textbf{response generator} ($\S$\ref{dgs}) intentionally fine-tuned for response generation, and \textbf{response selection} ($\S$\ref{rs}).

Given a dialogue history, $\{u^T_1, u^R_2, u^T_3, u^R_4, ..., u^R_{t-1}, u^T_{t}\}$, ending with the last utterance spoken by the client in a dialogue between a client and a counselor, the motivation behind a client-centric dialogue system is to \textit{accurately identify the client's behavior} and \textit{select the appropriate counseling strategy}. Here, $T$ and $R$ refer to the client and counselor, respectively, derived from the last characters in the words \verb+client+ and \verb+counselor+. For brevity, this convention will be consistently used thereafter and not repeated.

Therefore, to comprehend both clients' behaviors and counselors' strategies in text-based counseling conversations, we propose utilizing the publicly available counseling conversational dataset, \verb+Xinling+ \cite{li2023understanding}, which is annotated with rich labels containing information on clients' behaviors and counselors' strategies. For detailed definitions of these labels, please refer to the original paper \cite{li2023understanding}.

To enhance the practicality of the response generator for mental health support, we advocate a two-stage fine-tuning approach, considering the limited availability of actual counseling dialogues. In the initial phase, we propose using a large-scale, diverse multi-turn dialogue dataset, \texttt{SmileChat} \cite{qiu2023smile}, which is publicly accessible, for warm-up parameter-efficient fine-tuning. Subsequently, to better align with real-world application scenarios, we employ the \texttt{Xinling} dataset, consisting of authentic dialogues, for the second-stage downstream parameter-efficient fine-tuning.

Note that the classifier for client behavior recognition can be used to label the client's behaviors in \texttt{SmileChat}. However, an auxiliary task is required to label the counselor's strategies in \texttt{SmileChat}. Therefore, we introduce an auxiliary task: \textbf{counselor strategy recognition}, which is illustrated in $\S$\ref{csr}.

\subsection{Client Behavior Recognition}\label{cbr}
Considering a dialogue history ending with the last utterance spoken by the client between a client and a counselor, as illustrated in Equation \ref{eq:dialogue-history}, the goal is to accurately identify the client's behaviors.

\begin{equation}\label{eq:dialogue-history}
    D_h=\left\{u^T_1, u^R_2, u^T_3, u^R_4, ..., u^R_{t-1}, u^T_{t} \right\}
\end{equation}

Therefore, the dialogue context can be formulated as follows:
\begin{equation}\label{eq:dialogue-context}
    D_c=\{u^T_1, u^R_2, u^T_3, u^R_4, ..., u^R_{t-1}\}
\end{equation}

To construct training, validation, and test sets, each sample is represented as $(x_i, Y_i)\in \mathcal{D_{\mathrm{client}}}$, where $Y_i$ represents the subset in the label space of clients' behaviors introduced in \texttt{Xinling}. In reality, there are possibly multiple sentences in the client's utterance, and each sentence accordingly maps to a single label. To simplify the classification task, we restructure the clients' lengthy utterances, originally labeled with multiple categories, into pairs of sentences and corresponding labels. Specifically, given an $i$-th sentence in the client's utterance, we can obtain $(s_{i}^T, y_i)$, where $s_{i}^T\in u_t^T$ and $y_i$ is the label of clients' behaviors. Thus, we represent the input for client behavior prediction as follows:

\begin{equation}
    x_i=[D_c; \mathrm{[SEP]}; s^T_{i}]
\end{equation}
where $\mathrm[;]$ denotes the operation of textual concatenation.

\begin{figure}[t!]
    \centering
    \includegraphics[width=6.2cm]{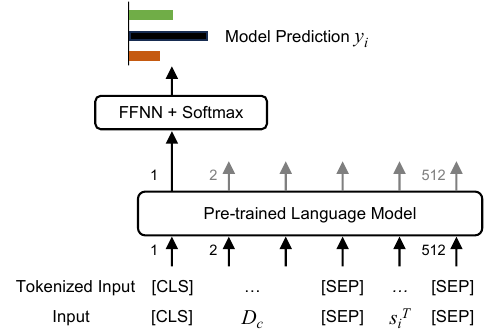}
    \caption{Mechanism of client behavior recognition.}
    \label{fig:cbr}
\end{figure}

The mechanism of client behavior recognition is presented in Figure \ref{fig:cbr}. To facilitate a dialogue system in understanding the client's states, we train a fully-connected feed-forward neural network (FFNN) with a softmax activation function to identify clients' behaviors based on a pre-trained language model.

\subsection{Counselor Strategy Selection}\label{css}
A dialogue session with a golden response spoken by the counselor can be formulated as follows:

\begin{equation}
    D_s=\{u^T_1, u^R_2, u^T_3, u^R_4, ..., u^R_{t-1}, u^T_{t}, R_g\}
\end{equation}
where $R_g$ represents the ground truth (a.k.a golden response) spoken by the counselor.

To recap, strategies behind a counselor's response contain explicit strategies with specific label descriptions and implicit strategies hidden in the dialogue session. If we use a dialogue history to predict the counselor's strategies, it may seem contradictory to the task of recognizing client behavior. Motivated by the notion that \textit{similar problems often have analogous solutions}, we employ dense retrieval to address these challenges. This approach facilitates the dialogue system in determining the most appropriate strategy by identifying $j$ semantically similar samples during both the training and inference processes. For simplicity, we set $j$ to 1.

Specifically, to obtain the most similar dialogue session, we need to build a dialogue retrieval base $\mathcal{D}_{\mathrm{base}}$. For each dialogue session, $D_s\in \mathcal{D}_{\mathrm{base}}$, we first split it into two parts: the dialogue history $D_h$ and the ground truth $R_g$. We then construct mapping pairs $\{S_h, R_g\}$ between dialogue history $D_h$ and ground truth $R_g$, where the string of a dialogue history is denoted as $S_h=[u^T_1; u^R_2; u^T_3, u^R_4; ...; u^R_{t-1}; u^T_{t}]$. Therefore, when considering a brand-new dialogue history $\hat{D_h}$, we propose to utilize the embedding model and apply dense retrieval to find a sample with the most minimal distance.

In this paper, we utilize the Euclidean distance as the metric, which is widely adopted for similarity measurement.

\begin{equation}
    d(\hat{D_h}, D_h) = \sqrt{\sum_{i=1}^{n}(q_i-p_i)^2}
\end{equation}
where $q$ and $p$ denote the mapping vectors of $\hat{D_h}$ and $D_h$ in Euclidean $n$-space, respectively.

\paragraph{Training Process} During the training process, since we have access to the ground truth, we directly utilize the counselor's strategies.

\paragraph{Inference Process} However, during the inference process, as we cannot obtain the ground truth, we adopt the counselor's strategies from the retrieved samples for response generation.

\subsection{Input Packer}\label{ip}
\begin{figure}[t!]
    \centering
    \includegraphics[width=8.8cm]{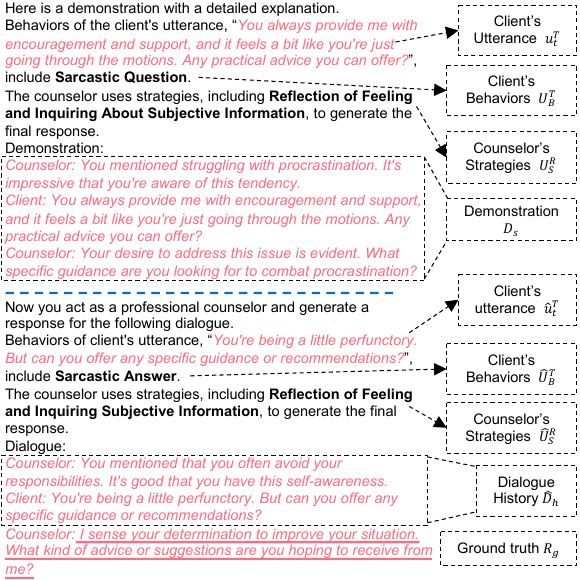}
    \caption{Template of the input packer. During the fine-tuning process, the ground truth $R_g$ is treated as a supervised signal. In the inference process, our goal is to generate the response.}
    \label{fig:ip}
\end{figure}

To better harness the inherent inference ability of large language models, we redefine the conventional dialogue generation task using the instruction-following paradigm and address it through parameter-efficient fine-tuning. The template of the input packer is illustrated in Figure \ref{fig:ip}, comprising two main sections: a demonstration part and a brand-new dialogue part, separated by a dashed blue line. Each section consists of instructions and placeholders.

\subsubsection{Instructions} The instructions serve to provide the model with a well-defined role and precise task details for dialogue generation. For the dialogue generation task, the text that is non-bold and black is our instructions, as shown in Figure \ref{fig:ip}.

\subsubsection{Placeholders}
Based on the provided dialogue history $\hat{D}_h$, which includes the client's utterance $\hat{u}_t^T$ and its context $\hat{D}_c$ in the brand-new dialogue segment, our first step is to predict the client's current behaviors $\hat{U}_B^T$ using the method outlined in $\S$\ref{cbr}.

Furthermore, we obtain the closest dialogue sample using the method described in $\S$\ref{css}. Hence, we select the closest sample as the demonstration $D_s$. Accordingly, we obtain the client's utterance $u_t^T$, the client's behaviors $U_B^T$, and the counselor's strategies $U_S^R$ for the demonstration part. During the interference process, specifically in the brand-new dialogue part, the counselor's strategies $\hat{U}_S^R$ remain the same as those in $U_S^R$.

\subsubsection{Packer Function} To summarize, the client input undergoes processing through the packer function, as illustrated in Figure \ref{fig:ip}, before reaching the response generator. This process is defined as follows:
\begin{equation}
    I_t=\mathbf{P}(u_t^T, U_B^T, U_S^R, D_s, \hat{u}_t^T, \hat{U}_B^T, \hat{U}_S^R, \hat{D}_h)
\end{equation}
where $\mathbf{P}(.)$ is the packer function used to format all input elements into a string.

\subsection{Response Generator}\label{dgs}
\subsubsection{Model} ChatGLM2-6B\footnote{https://huggingface.co/THUDM/chatglm2-6b} \cite{zeng2022glm} is an open-source bilingual (Chinese-English) chat model, which is trained with a context length of 8192 during dialogue alignment, enabling more turns in conversation. Therefore, we choose it as the foundational model for our response generator.

\subsubsection{Dataset Labeling}
To adapt to the real-world downstream task, we fine-tune classifiers using the dialogue dataset, \texttt{Xingling}, to annotate both the client's and counselor's utterances, obtaining pseudo-labels for the dialogue dataset, \texttt{SmileChat}. The mechanism of fine-tuning the classifier to identify the behaviors of clients' utterances and strategies of counselors' utterances is illustrated in $\S$\ref{cbr} and $\S$\ref{csr}, respectively.

\subsubsection{Dialogue Generation} Considering our collected and processed dataset $\mathcal{D_{\mathrm{dial}}}=\{ D_1, D_2, ..., D_i, ..., D_n \}$, where each $D_i$ represents a single multi-turn dialogue, we split each dialogue into multiple training sessions to train a response generator. Specifically, for a sampled $t$-turn dialogue with the ground truth, $d=\{u^T_1, u^R_2, u^T_3, u^R_4, ..., u^T_t, R_g\} \sim \mathcal{D_{\mathrm{dial}}}$, we build a dialogue model that can predict the counselor's utterance $R_{g}$ given the dialogue history $D_h$. We adopt the input packer in $\S$\ref{ip} to construct the training data $\{I_t, R_g\}$. For both stages, the demonstration is retrieved from the validation set in the dataset, \texttt{Xinling}. Our objective is to maximize the likelihood probability as follows:

\begin{eqnarray}
\mathbb{E}_{d \sim \mathcal{D_{\mathrm{dial}}}}\prod_{l=1}^{L} \mathbb{P}(R_g|I_t)
\end{eqnarray}
where $L$ is the sequence length of the ground truth $R_g$.

\subsubsection{Response Generation} Moving forward, we input $I_t$ into the response generator, which has been trained through a two-stage fine-tuning process, to generate 10 response candidates.

\subsection{Response Selection}\label{rs}
Due to the uncertainty and diversity inherent in model generation, we propose adopting the sample-and-rank paradigm to select the optimal response. To achieve this objective, we suggest employing the widely used response selection architecture: the Cross-encoder. This architecture facilitates rich interactions between the dialogue history $D_h$ and response candidate $R_i$. We consider the first output of the pre-trained language model as the history-candidate embedding.

\begin{equation}
    y_{D_h, R_i} = h_1 = first(Tr(D_h, R_i))
\end{equation}
Where $first$ is the function that takes the first vector from the sequence of vectors produced by the pre-trained language model $Tr$.

To score a candidate, a linear layer, denoted as $W$, is applied to the embedding $y_{D_h, R_i}$ to transform it from a vector to a scalar: $s(D_h, R_i) = y_{D_h, R_i}W$. The neural network is trained to minimize cross-entropy loss, where the logits are $s(D_h, R_1), ..., s(D_h, R_{10})$. Here, $R_1$ represents the ground truth, and the rest are negative samples randomly selected from the in-batch set.

\subsection{Counselor Strategy Recognition}\label{csr}
A dialogue session with a response from the counselor can be formulated as $D_s=\{D_h, R_g\}$. Referring to the $\S$\ref{cbr}, we represent the input for counselor strategy prediction as $x_i=[D_h; \mathrm{[SEP]}; s^R_{i}]$, where $s^R_i$ is the $i$-th sentence in the ground truth $R_g$.

\begin{table}[t!]
\centering
\caption{Data statistics of both client behavior and counselor strategy prediction.}
\label{tab:data-statistics}
\scalebox{0.8}{
\begin{tabular}{c|c|c|c}
\toprule
\textbf{Task}      & \textbf{\# Training} & \textbf{\# Validation} & \textbf{\# Test} \\\midrule
Client Behavior Prediction    & 18824 & 3683       & 3627 \\
Counselor Strategy Prediction & 21082 & 4272       & 4045\\
\bottomrule
\end{tabular}
}
\end{table}

\section{Experiments}
All experiments are performed using NVIDIA A100 8$\times$80G GPUs. For the tasks of client behavior recognition, counselor strategy recognition, and response selection, we prepend a speaker token (\verb+[client]+ or \verb+[counselor]+) to each utterance to identify the speaker.

\subsection{Client Behavior Recognition}

\subsubsection{Data Statistics}
After processing the data, we present the statistics for client behavior prediction in Table \ref{tab:data-statistics}.

\subsubsection{Hyperparameters}
We employ a pre-trained Chinese RoBERTa-large \cite{cui2020revisiting} model\footnote{https://huggingface.co/hfl/chinese-roberta-wwm-ext-large}, commonly used for text classification. The hyperparameters used for fine-tuning the model in client behavior recognition are provided in Table \ref{tab:hyperparameters-for-finetune}.

\begin{table}[t!]
\caption{Hyperparameters for fine-tuning the model used for both client behavior and counselor strategy recognition.}
\label{tab:hyperparameters-for-finetune}
\centering
\scalebox{0.8}{
    \begin{tabular}{c|c}
    \toprule
    \textbf{Hyperparameters} & \textbf{Values} \\ \midrule
    epochs & 10 \\\midrule
    batch size & 16 \\\midrule
    learning rate (lr) & 2e-5 \\\midrule
    weight decay ($\lambda$) & 0.01 \\ \bottomrule
    \end{tabular}
    \begin{tabular}{c|c}
    \toprule
    \textbf{Hyperparameters} & \textbf{Values} \\ \midrule
    seed number & [42, 43, 44, 45] \\\midrule
    warmup ratio & 0.1 \\\midrule
    momentum values [$\beta_1, \beta_2$] & [0.9, 0.999] \\\midrule
    dropout rate & 0.1 \\ \bottomrule
    \end{tabular}
}
\end{table}

\subsubsection{Results}
The best checkpoint with the highest accuracy in the validation set is retained for each seed. The results of accuracy among the four seeds in the test set are reported in Table \ref{tab:accuracy}, demonstrating comparable accuracy to the original paper. Therefore, the checkpoint trained with the seed of 42 is selected for client behavior recognition in our paper.

\begin{table}[t!]
\centering
\caption{Results of client behavior and counselor strategy prediction in the test set, as well as response selection.}
\label{tab:accuracy}
\scalebox{0.8}{
\begin{tabular}{c|c|c|c|c}
\toprule
\textbf{Task}      & \textbf{42 (\%)}    & \textbf{43 (\%)}    & \textbf{44 (\%)}    & \textbf{45 (\%)}    \\\midrule
Client Behavior Prediction    & \textbf{85.97} & 85.47 & 84.97 & 85.31 \\
Counselor Strategy Prediction & 80.15 & \textbf{80.77} & 80.25 & 80.67  \\
Response Selection (R@1/10) & 81.84 & \textbf{82.10} & 81.77 & 81.54  \\
\bottomrule
\end{tabular}
}
\end{table}

\subsection{Counselor Strategy Selection}
We utilize the publicly available embedding model \texttt{BAAI/bge-large-zh-v1.5}\footnote{https://huggingface.co/BAAI/bge-large-zh-v1.5}, accessible on Hugging Face. The demonstrations used to construct the training and test sets for response generation are retrieved from the validation set in \texttt{Xinling}, as illustrated in the second row of Table \ref{tab:data-statistics-for-fine-tuning}. Furthermore, during the inference process, the counselor's strategy is also retrieved from the validation set in \verb+Xinling+.

\subsection{Input Packer}
We provide data statistics for fine-tuning the response generator in Table \ref{tab:data-statistics-for-fine-tuning}.

\begin{table}[t!]
\centering
\caption{Data statistics. The datasets represented by $\diamondsuit$ and $\clubsuit$ are utilized for response generator and response selection, respectively.}
\label{tab:data-statistics-for-fine-tuning}
\scalebox{0.8}{
\begin{tabular}{c|c|c|c}
\toprule
\textbf{Data Type}      & \textbf{\# Training} & \textbf{\# Validation} & \textbf{\# Test} \\\midrule
SmileChat$^{\diamondsuit}$ & 310087 & -       & -\\
Xingling$^{\diamondsuit}$    & 15850 & \underline{3150} (only used for retrieval)      & 3072 \\
Xingling$^{\clubsuit}$    & 15850 & 3150       & 3072 \\
\bottomrule
\end{tabular}
}
\end{table}

\subsection{Response Generator}
\subsubsection{Parameter-efficient Fine-tuning}
We apply the Low-Rank Adaption (LoRA \cite{hu2021lora}) to all linear layers in the ChatGLM2-6B model for efficient fine-tuning. We present the hyperparameters for fine-tuning the response generator in Table \ref{tab:hyper-parameters}.

\begin{table}[t!]
\caption{Parameters of parameter-efficient fine-tuning.}
\label{tab:hyper-parameters}
\centering
\scalebox{0.8}{
    \begin{tabular}{c|c|c|c|c|c|c}
    \toprule \textbf{Epoch} & \textbf{\begin{tabular}[c]{@{}l@{}}Learning\\ Rate\end{tabular}} & \textbf{\begin{tabular}[c]{@{}l@{}}Batch\\ Size\end{tabular}} & \textbf{\begin{tabular}[c]{@{}l@{}}LoRA\\ Rank\end{tabular}} & \textbf{\begin{tabular}[c]{@{}l@{}}LoRA\\Dropout\end{tabular}} & \textbf{\begin{tabular}[c]{@{}l@{}}LoRA\\$\alpha$\end{tabular}} & \textbf{Seed} \\ \midrule
    2 & 1e-4 & 1 & 16 & 0.1 & 64 & 1234 \\
    \bottomrule
    \end{tabular}
}
\end{table}

\subsubsection{Dialogue Generation}  During the generation process, we set the maximum sequence length to 8192, the temperature to 0.8, and \texttt{top\_p} to 0.8. Finally, we obtain 10 responses and select the optimal one using our trained response selector.

\subsection{Response Selection}
For the evaluation metric, we measure Recall@$k$/$N$, where each test example has $N$ possible response candidates to select from, abbreviated as R@$k$/$N$. Since our objective is to select the optimal response from these candidates, we set $k$ to 1 and $N$ to 10.

\subsubsection{Data Statistics}
We present the data statistics for response selection in Table \ref{tab:data-statistics-for-fine-tuning}.

\subsubsection{Hyperparameters}
We utilize the hyperparameters for response selection in Table \ref{tab:hyperparameters-for-finetune}. One notable difference is that we set the batch size to 1, considering that each batch consists of one ground truth and nine negatives.

\subsubsection{Results}
The results of response selection among the four seeds in the test set are presented in Table \ref{tab:accuracy}. Consequently, we choose the checkpoint trained with seed 43 for response selection in our paper.

\subsection{Counselor Strategy Recognition}
\subsubsection{Data Statistics}
We provide the data statistics for counselor strategy prediction in Table \ref{tab:data-statistics}.

\subsubsection{Hyperparameters}
We present the hyperparameters used for fine-tuning the model in counselor strategy recognition in Table \ref{tab:hyperparameters-for-finetune}.

\subsubsection{Results}
We present the accuracy results for the four seeds in the test set, as depicted in Table \ref{tab:accuracy}. Consequently, we select the checkpoint trained with seed 43 for data annotation in our paper.

\section{Results}

\subsection{Automatic Evaluation}
\subsubsection{Metrics}
To conduct automatic evaluation, we utilize the following evaluation metrics: Perplexity (PPL) \cite{jelinek1977perplexity}, METEOR \cite{banerjee2005meteor}, BLEU-1/2/3 \cite{papineni2002bleu}, Rouge-L \cite{lin2004rouge}, and Distinct-1/2 (D-1/2) \cite{li2015diversity}.

\subsubsection{Results}
We present the results of the automatic evaluation in Table \ref{tab:automatic-evaluation}. All metrics for automatic evaluation indicate enhanced performance in the real-life test set, highlighting the effectiveness and practicality of our proposed client-centric dialogue system.

\begin{table}[t!]
\centering
\caption{Results of automatic evaluation in the test set.}
\scalebox{0.7}{
\begin{tabular}{c|c|c|c|c}
\toprule
              & \textbf{PPL} ($\Downarrow$)        & \textbf{METEOR} ($\Uparrow$)       & \textbf{BLEU-1} ($\Uparrow$)       & \textbf{BLEU-2} ($\Uparrow$)            \\\midrule
Baseline & 3.39          & 16.8          & 9.1          & 4.1                    \\\midrule
Fine-tuning  & \textbf{1.26} & \textbf{22.9} & \textbf{24.1} & \textbf{11.6} \\
\toprule
\toprule
&  \textbf{BLEU-3} ($\Uparrow$)      & \textbf{Rouge-L} ($\Uparrow$)      & \textbf{D-1} ($\Uparrow$)          & \textbf{D-2} ($\Uparrow$)          \\\midrule
Baseline           & 1.5          & 12.4          & 62.1          & 87.8          \\\midrule
Fine-tuning   & \textbf{5.3} & \textbf{27.6} & \textbf{86.6} & \textbf{97.2} \\\bottomrule
\end{tabular}
}
\label{tab:automatic-evaluation}
\end{table}

\subsection{Human Evaluation}
For human evaluation, we recruit three professional counselors, each with more than two years of counseling experience. Therefore, we sample 100 examples, each comprising a dialogue history with three responses: the ground truth, the response generated by the baseline model (without fine-tuning), and the response generated by our trained dialogue system. The last two responses are selected from 10 candidates by our response selector, respectively. Three counselors are tasked with selecting the optimal response from three shuffled options. They consider which one is more suitable for the dialogue history from a holistic perspective.

We present the results of the human evaluation in Table \ref{tab:human-evaluation}. Each annotator chooses a model response. The chosen model scores 1 point, and the other two models score 0 points. Thus, we can obtain three arrays of 300-dimensional scores and calculate the significant difference between each pair of models separately. The results show that the $p$-value between the fine-tuned model and the ground truth is 0.04, while the $p$-values for the other two comparisons are 0.00 (Student's t-test, $^{**}p=0.01$).

The average selection times are 50, 42, and 8 for ground truth, fine-tuning, and baseline, respectively. Three professional counselors generally agree that the quality of responses generated by our trained dialogue system is comparable to that of the ground truth. Therefore, we conclude that the quality of responses generated by our proposed dialogue system is comparable to the ground truth, demonstrating the practicality and effectiveness of our client-centric dialogue system.

\begin{table}[t!]
\centering
\caption{Results of human evaluation of 100 samples. Each annotator chooses an optimal model response among three models: baseline, fine-tuned, and ground truth.}
\label{tab:human-evaluation}
\scalebox{0.8}{
\begin{tabular}{c|c|c|c}
\toprule
\textbf{Annotator}      & \textbf{Baseline}     & \textbf{Fine-tuning}    & \textbf{Ground Truth}    \\\midrule
Annotator A & 3 & 45 & 52  \\
Annotator B & 10 & 44 & 46 \\
Annotator C & 12 & 36 & 52 \\\hline
Average & 8 & 42 & 50 \\
\bottomrule
\end{tabular}
}
\end{table}

\begin{figure}[t!]
    \centering
    \includegraphics[width=0.93\columnwidth]{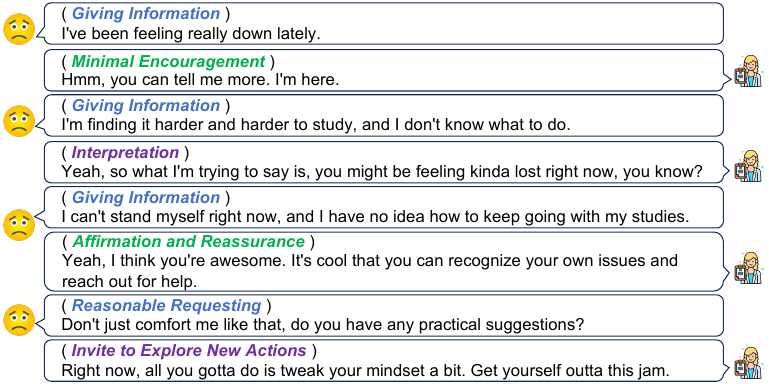}
    \caption{An illustration of a case study of our client-centric dialogue system.}
    \label{fig:case-study}
\end{figure}

\subsection{Case Study}
We present an illustration of a case study of our client-centric dialogue system in Figure \ref{fig:case-study}, demonstrating that it can better consider the client's behavior. With its ability to select an appropriate counseling strategy and with the assistance of response selection, our system can provide better responses in client-machine interaction scenarios.

\section{Conclusion}
In conclusion, the client-centric dialogue system for mental health support represents a viable approach for enhancing client well-being, addressing issues of accessibility, affordability, and stigma. Through both automatic and human evaluations, our system has demonstrated its effectiveness and practicality in real-life mental health support scenarios. Notably, in a simulated client-virtual-counselor interaction, the system successfully predicted client behaviors, selected appropriate counselor strategies, and generated accurate and suitable responses with the assistance of response selection.

\bibliographystyle{IEEEtran}
\bibliography{refs}

\end{document}